\documentclass{article}

\usepackage[preprint]{neurips_2020}

\usepackage[utf8]{inputenc} 
\usepackage[T1]{fontenc}    
\usepackage{hyperref}       
\usepackage{url}            
\usepackage{booktabs}       
\usepackage{multirow}
\usepackage{amsfonts}       
\usepackage{nicefrac}       
\usepackage{microtype}      
\usepackage{adjustbox}
\usepackage{bm}
\usepackage{graphicx}
\usepackage{dsfont}
\usepackage{tikz}
\usepackage{amsmath}
\usepackage{authblk}
\title{GraphCL: Contrastive Self-Supervised Learning of Graph Representations}

\author[1,2]{Hakim Hafidi}
\author[1]{Mounir Ghogho}
\author[2]{Philippe Ciblat}
\author[3]{Ananthram Swami}
\affil[1]{TICLab, College of Engineering and Architecture, Universit\'e Internationale de Rabat,   Morocco \authorcr
  \{\tt hakim.hafidi, mounir.ghogho\}@uir.ac.ma}
\affil[2]{LTCI, Communications and Electronics Department, Telecom ParisTech, Institut Polytechnique de Paris, France\authorcr
  \tt philippe.ciblat@telecom-paris.fr}
\affil[3]{United States Army Research Laboratory, Adelphi, Maryland, USA \authorcr
  \tt ananthram.swami.civ@mail.mil}

\begin{document}

\maketitle

\begin{abstract}
We propose Graph Contrastive Learning (GraphCL), a general framework for learning node representations in a self supervised manner. GraphCL learns node embeddings by maximizing the similarity between the representations of two randomly perturbed versions of the intrinsic features and link structure of the same node's local subgraph. We use graph neural networks to produce two representations of the same node and leverage a contrastive learning loss to maximize agreement between them. In both transductive and inductive learning setups, we demonstrate that our approach significantly outperforms the state-of-the-art in unsupervised learning on a number of node classification benchmarks.  
\end{abstract}

\section{Introduction}

In many fields, the rapidly increasing volume and complexity of data hinders actionable insights. Graphs offer an unifying framework for aligning structured and unstructured data. However, graphs have long been poorly exploited because of their complexity, and limited approaches in dealing with content associated with nodes and links. Recently, graph representation learning has attracted the attention of the scientific community as a way of analysing graphs and helping to leverage the richness of information that resides in unstructured data. Graphs are characterized by a set of nodes, which represent the entities, and a set of links connecting them. Nodes may be of different types, and may further be associated with several features. And links may represent different relationships and may also be associated with different attributes or semantic content. One of the major challenges facing graph representation learning is learning node embeddings which capture both node features and the graph structure. These representations can then be fed into downstream machine learning models.

Most successful approaches for graph representation have been great efforts to generalize neural networks to graph data and fall under the umbrella of Graph Neural Networks (GNNs) or Deep Geometric Learning \citep{atwood2016diffusion,kipf2016semi,bronstein2017geometric,xu2018powerful}. These approaches have achieved remarkable results in a number of important tasks such as node classification \citep{hamilton2017inductive,chami2019hyperbolic,luan2019break} and link prediction \citep{kipf2016variational,zhang2018link}. However, these methods are very reliant on human intervention and suffer from the necessity of some form of supervision. This requires a high cost, expert knowledge in the domain and the use of annotated data, which is not often available. Thus, the need to develop methods capable of learning representations in an unsupervised manner is essential.

\par
In order to compensate for the absence of labels or predefined tasks, some unsupervised methods have adopted the homophily hypothesis, which states that linked nodes should be adjacent in the embedding space \citep{hoff2002latent}. Inspired by the Skipgram algorithm for embedding words into a latent space, where adjacent vectors correspond to co-occurring words in a sentence \citep{mikolov2013efficient}, a majority of these methods use random walks to generate sentence-like sequences where co-occurring nodes are close in the embedding space \citep{perozzi2014deepwalk,grover2016node2vec}. Other methods such as autoencoders, also employ the homophily hypothesis by reconstructing either the adjacency matrix or the neighborhood of a node \citep{wang2016structural,kipf2016variational}. Despite their success in learning relatively powerful representations, relying on the homophily hypothesis biases these methods towards emphasizing the direct proximity of nodes over topological information \citep{wang2016structural}. More recently, \citep{velivckovic2018deep} proposed Deep Graph Infomax (DGI) that learns representations by training a discriminator to distinguish between representations of nodes that belong to the graph from nodes coming from a corrupted graph. Leveraging recent advances in unsupervised visual representations \citep{hjelm2018learning}, the success of  DGI has been attributed to the maximization of mutual information between global and local parts of the input. This requires learning global representations of the entire graph which can be very costly and even intractable when dealing with large graphs.
\par

In this work, we introduce GraphCL, a general contrastive learning framework that learns node embeddings by maximizing agreement/similarity between the representations of two randomly perturbed versions of the same node's local subgraph. In addition to learning node representations that are robust to random perturbations of the graph, GraphCL allows for an efficient self-supervised learning of node representations. 
\par
GraphCL is inspired by the success of a recent approach which leverages contrastive learning losses to learn visual representations that capture shared information across multiple views of the same image. These methods are based on the assumption that \textit{important} information is shared between different views of the world. The authors of \citet{chen2020simple} use data augmentation techniques to generate multiple views of the same image, while \citet{tian2019contrastive} consider different channels of an image as different views.

In GraphCL, for each node, random perturbations are applied to its $L$-hop subgraph. The perturbation consists of randomly dropping a subset of edges and nodes' intrinsic features of its $L$-hop subraph. The dropout probabilities are hyperparameters in the learning process. 

We show that GraphCL achieves a new state-of-the-art in unsupervised node representation by demonstrating how it consistently outperforms previous state-of-the-art methods on both transductive and inductive setups.

\section{Related work}
Most unsupervised graph representation learning methods can be described as contrastive approaches. Their main objective is to train an encoder to be contrastive between pairs of samples that follow observations in the data and thus capture statistical dependencies of interest (a.k.a positive examples) and those that are not (a.k.a negative examples). Contrastive approaches have had great success for learning words representations \citep{mikolov2013efficient,collobert2008unified}. They have also been used for learning visual representations dating back to \citep{hadsell2006dimensionality}, and have started to show promising results in recent works \citep{misra2019self,zhuang2019local}, achieving competitive results when compared to strong supervised baselines \citep{chen2020simple}. Researchers have extended many of these methods to learn representations of graph structured data.  DeepWalk \citep{perozzi2014deepwalk}  and  Node2vec \citep{grover2016node2vec}  are inspired by language models such as Word2vec \citep{mikolov2013efficient}. DGI \citep{velivckovic2018deep} adapted ideas from Deep InfoMax  \citep{hjelm2018learning} method that learns representations of high-dimensional data. GraphCL is also contrastive in this sense, as we learn a classifier to distinguish between positive examples (i.e., pairs of augmented views of the same node) and representations of the other nodes. 
\par
Contrastive methods differ in the choice of different components such as a sampling strategy to select positive and negative examples and an encoder to embed a data sample into a destination space. The above mentioned methods on graph representation learning often take into account the structure of the graph; positive and negative examples correspond to adjacent and distant nodes respectively. DeepWalk and Node2vec, for example, use different policies to generate fixed length random walks to find positive examples whereas random nodes correspond to negative examples. Both methods use a lookup table as an encoder. Graph Autoencoder encourages the use of first order neighbors as positive examples and use Graph Convolutional Networks as encoders \citep{kipf2016variational}, while  \citep{bojchevski2017deep} use higher order neighbors as positive examples and use a Multi-Layer Perceptron to encode node features. DGI employs a different strategy:  node representations obtained from the graph correspond to positive examples, while negative examples correspond to embeddings of a corrupted graph. DGI also uses different architectures of Graph Convolutional Networks as encoders \citep{kipf2016semi,hamilton2017inductive}.
\par
Graph convolution Networks enforce an inductive bias that adjacent nodes have similar representations.  Instead of explicitly injecting the graph structure information to sample positive and negative examples, we leverage the latest insights about GCN encoders and employ a strategy that is completely different  from the methods discussed above. We get two different representations of each node by randomly sampling two subgraphs around the node. These two representations will correspond to a pair of positive examples.

\section{Methodology}
In this section, we define some needed notation, formulate the learning problem of interest, and then provide details of the proposed solution. 

\subsection{Background}

\paragraph{Problem formulation.}
Let $\mathcal{G=(V,E)}$ be an undirected graph where $\mathcal{V}$ is a set of nodes and $\mathcal{E} \subseteq \mathcal{V} \times \mathcal{V}$ is a set of edges. Each node $u \in \mathcal{V}$ is represented by a feature vector $x_{u} \in \mathbb{R}^{P}$. An adjacency matrix $A \in \mathbb{R}^{N \times N}$ represents the topological structure of the graph where $N = |\mathcal{V}|$ is the number of nodes in the graph. Without loss of generality we assume the graphs to be unweighted i.e $A_{u,v}=1$ if $(u,v) \in \mathcal{E} $ and $A_{u,v}=0$ otherwise. We are also provided with $ X = \{x_{1},x_{2},\dots , x_{N}\}$, a set of nodes' features.
\par
Let $\mathcal{G}_{u}=(\mathcal{V}_{u},\mathcal{E}_{u})$ be the $L-$hop subgraph centred at node $u$ and $X_{u} = \{x_{j}\}_{j\in \mathcal{V}_{u}}$ be the set of feature vectors corresponding to the nodes in $u$'s neighborhood subgraph. The relational information within $u$'s subgraph is represented by its corresponding adjacency matrix $A_{u}$.
Our objective is to learn an effective representation of nodes without human intervention. This will be done through the learning of a graph neural network encoder $f$ that maps both node level information and the graph structure to a higher level representation i.e $f(X_{u},A_{u}) = h_{u}^{(L)} \in \mathbb{R}^{P'}$ for each $u \in \mathcal{V}$, where $P'$ is the embedding size. It is worth pointing out that the embedding $h_{u}^{(L)}$ corresponds to the output of the $L-$th layer of the GNN, which involves the nodes of node $v$'s $L$-hop subgraph. In the remainder of the paper, $h_{u}$  refers to the output of the GNN's last layer, i.e.  $h_{u}=h_{u}^{(L)}$.    

\paragraph{Graph Neural Networks (GNNs).}
GNNs are a class of graph embedding architectures which use the graph structure in addition to node and edge features to generate a representation vector (i.e., embedding) for each each node. Recent GNNs learn node representations by aggregating the features of neighboring nodes and edges. The output of the $l$-th layer of these GNNs is generally expressed as
\begin{equation}
\label{gnn}
    h_{u}^{(l)}= COMBINE^{(l)}(h_{u}^{(l-1)},AGGREGATE^{(l)}(\{(h_{u}^{(l-1)},h_{v}^{(l-1)}): v \in \mathcal{N}(u)\})),
\end{equation}
where $h_{u}^{(l)}$ is the feature vector of node $u$ at the $l$-th layer initialized by $h_{u}^{(0)}=x_{u}$ and $\mathcal{N}(u)$ is the set of first-order neighbors of node $u$. Different GNNs use different formulations of the COMBINE and AGGREGATE functions; the ones used in this work are described in the next section.

\subsection{GraphCL}
\label{metho:Grpahcl}
\begin{figure}[t]
\centering
\begin{adjustbox}{width=\textwidth}

\input{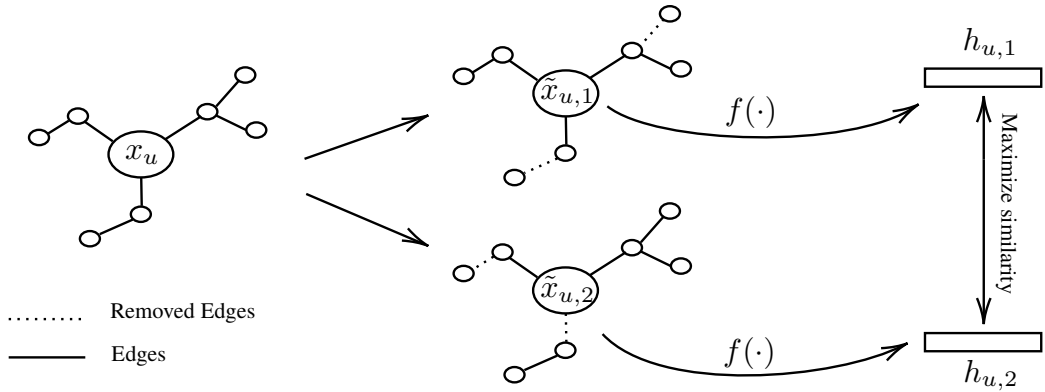}

\end{adjustbox}
\caption{A high-level overview of our method for a subgraph around node $u$. Refer to section \ref{overview} for details}
\label{fig:overview}
\end{figure}

GraphCL's objective is to learn node representations by maximizing the similarity between the embeddings of two randomly perturbed versions of the same node's neighborhood subgraph using a contrastive loss in the embedding space. This framework has three main components: a stochastic perturbation, a GNN based encoder and a contrastive loss function. We first introduce each of these components, and then give a high-level overview of the proposed method. 
\begin{itemize}
    \item \textbf{Stochastic perturbation.} We apply a stochastic perturbation to the $L$-hop subgraph centered at each node, that results in two neighborhood subgraphs that allow us to obtain two representations of the same node which we consider as positive examples. In this work we consider simultaneous transformation of both node features and the connectivity of the subgraph. The subgraph structure is transformed by randomly dropping edges with probability $p$ using samples from a Bernoulli distribution. For the node's intrinsic features, we apply a similar strategy by simply applying dropout to the input features; see illustration in Figure \ref{fig:overview}.
    \item \textbf{Graph neural network encoder.} We apply a GNN based encoder that learns representations of the two transformed $L$-hop subgraphs associated with each node $u$. Our framework supports several choices of GNN architectures. We choose the simple \textit{mean-pooling} propagation rule as introduced by \citep{hamilton2017inductive} as the main building block, and adopt different choices for the inductive and transductive setups. Details about the choices of architectures are given in section \ref{models}.
    \item \textbf{Contrastive loss function.} We define a \textit{pretext} prediction task that aims at identifying the corresponding positive example $h_{u,2}$ of a representation $h_{u,1}$ given a set of generated examples, with $h_{u,1}$ and $h_{u,2}$ being a positive pair of examples (i.e. obtained from the GNN representation of two transformations of the $L$-hop subgraph around the same node).
\end{itemize}

\par

We randomly sample a minibatch $\mathcal{B}$ containing $M$ nodes and define their corresponding $L$-hop subgraphs. We apply two transformations to each of the node's subgraphs resulting in $2M$ subgraphs enabling us to get positive pairs of representations for the contrastive prediction task. Instead of explicitly sampling  negative examples, we consider the other $(2M-2)$ examples within the minibatch as negative examples following the sampling strategy in \citep{chen2017sampling}.
\par
For each node $u$ in the minibatch, we compute the following loss function, which is based on a normalized temperature-scaled cross entropy:
\begin{equation}
    \label{loss1}
    l(u) = l_{1,2}(u) + l_{2,1}(u),
\end{equation}
where $l_{i,j}(u)$ is defined as
\begin{equation}
\label{lossv1}
    l_{i,j}(u)= - \log \frac{\exp(\mathrm{s}(h_{u,i}, h_{u,j})/\tau)}{\sum_{v\in \mathcal{B}} \mathds{1}_{[v \ne u]}\exp(\mathrm{s}(h_{u,i}, h_{v,i})/\tau)+\sum_{v \in \mathcal{B}} \exp(\mathrm{s}(h_{u,i}, h_{v,j})/\tau)},
\end{equation}
where $\mathrm{s}(h_{u,i}, h_{u,j})=h_{u,i}^{\top} h_{u,j}\slash\|h_{u,i}\|\|h_{u,j}\|$ is the cosine similarity between the two representations $h_{u,i}$ and $h_{u,j}$, $\mathds{1}_{[u \ne v]}$ is an indicator function equals to $1$ iff $u \ne v$ and $\tau$ a temperature parameter.

\subsection{Overview of GraphCL}
\label{overview}
For each sampled minibatch $\mathcal{B}$, we apply the following steps:
\begin{enumerate}
    \item For each node $u$ in the minibatch we define $(X_{u},A_{u})$ as the subgraph containing all nodes and edges that are at most $L$-hops from $u$ in the graph and their corresponding features;
    \item Draw two stochastic perturbations $t_{1}$ and $t_{2}$ as defined in section \ref{metho:Grpahcl} and apply them to $u$'s $L$-hop neighborhood subgraph:
    \begin{itemize}
        \item $(\widetilde{X}_{u,1},\widetilde{A}_{u,1})\sim t_{1}({X}_{u},{A}_{u})$
        \item $(\widetilde{X}_{u,2},\widetilde{A}_{u,2})\sim t_{2}({X}_{u},{A}_{u})$
    \end{itemize}
     
    \item Apply the encoder to the two representations of node $u$:
    \begin{itemize}
        \item $h_{u,1}=f(\widetilde{X}_{u,1},\widetilde{A}_{u,1}) $
        \item $h_{u,2}=f(\widetilde{X}_{u,2},\widetilde{A}_{u,2}) $
    \end{itemize}
    \item Update parameters of the encode, $f$ using the following loss function
    \begin{equation}
    \label{eq:loss2}
    \mathcal{L}= \frac{1}{|\mathcal{B}|} \sum_{u \in \mathcal{B}}  l(u),
    \end{equation}
    where $l(\cdot)$   is defined in equation Eq. (\ref{loss1}) 
\end{enumerate}

\section{Experiments}
We evaluate the effectiveness of GraphCL representations on both tranductive and inductive learning setups. The transductive learning setup consists of embedding nodes from a fixed graph (i.e. All nodes' features and the entire graph structure are known during the training time). On the other hand, the inductive learning setup consists of generating representation of unseen nodes or new graphs.
Following common practice, we opt for a linear evaluation of the learned node representations. Specifically, we use these representations to train a logistic regression model to solve multiclass node classification tasks on five well-know benchmark datasets, three for the transductive learning setup and two for the inductive setup. We summarize the datasets and the baselines respectively in sections \ref{sec:datsets} and \ref{sec:baselines}, provide model configuration and implementation details in section \ref{models}, and discuss the results in section \ref{results}.
 
\subsection{Datasets}
\label{sec:datsets}
For the transductive setting, we utilize Cora, Citeseer and Pubmed \citep{sen2008collective}, three citation networks where nodes are bag of words representations of documents and edges correspond to (undirected) citations. Each node belongs to one class. On the other hand, a protein-protein interaction dataset (PPI) is used for the the inductive setting on multiple graphs \citep{zitnik2017predicting}. It consists of multiple graphs corresponding to different human tissues where node features are the positional gene sets, motif gene sets and immunological signatures. Each node has several labels among 121 labels from the gene ontology. For the inductive setting on large graphs, we use a Reddit dataset \citep{hamilton2017inductive}. It represents a large social network where nodes correspond to Reddit posts (i.e. represented by their GloVe embedding \citep{pennington2014glove}) and edges connecting two posts mean that the same user commented on them. Labels are the posts' \textit{subreddit} and the objective is to predict the community structure of the social network. Statistics of the datasets are given in table \ref{data}.

\begin{table}[t]
\caption{Description of the datasets}
\label{data}
\centering
\begin{adjustbox}{width={\textwidth},totalheight={\textheight},keepaspectratio}
\begin{tabular}{@{}ccccccc@{}}
\toprule
\textbf{Dataset} & \textbf{Task} & \textbf{Nodes} & \textbf{Edges} & \textbf{Features} & \textbf{Classes} & \textbf{Train\slash Val\slash Test Nodes  } \\ \midrule
\textbf{Cora} & Transductive  & 2,708 & 5,429 & 1,433 & 7 & 140\slash500\slash1,000         \\
\textbf{Citeseer} &  Transductive & 3,327 & 4,732 & 3,707 & 6 & 120\slash500\slash1,000\\
\textbf{Pubmed} &   Transductive & 19,717 & 44,338 & 500 & 3 & 60\slash500\slash1,000\\
\textbf{Reddit} &   Inductive & 231,443 & 11,606,919 & 602 & 41 & 151,708\slash23,699\slash55,334\\
\multirow{2}{*}{\textbf{PPI}} &    \multirow{2}{*}{Inductive}  & 56,944 & \multirow{2}{*}{818,716} & \multirow{2}{*}{50} & 121 &   44,906\slash6,154\slash5,524 \\
  &    &   (24 graphs)  &     &   &(multilabel)    &  (20\slash2\slash2 graphs)      \\ \bottomrule
\end{tabular}
\end{adjustbox}
\end{table}

\subsection{Baselines}
\label{sec:baselines}
For the transductive learning tasks, we use four state-of-the-art unsupervised methods for comparison: Label Propagation (LP) \citep{zhu2003semi}, DeepWalk \citep{perozzi2014deepwalk},  Embedding Propagation (EP-B) \citep{duran2017learning}, and Deep Graph Infomax (DGI) \citep{velivckovic2018deep}. We also report the results of training logistic regression on the intrinsic input features only, and also on the concatenation of DeepWalk embeddings and the nodes' intrinsic features. Aside from unsupervised methods, we also compare our approach to strong supervised baselines, Graph Convolution Networks (GCN) \citep{kipf2016semi} and Graph Attention Networks (GAT) \citep{velivckovic2017graph}.
\par
For the inductive learning tasks, in addition to DeepWalk and DGI, we compare GraphCL with the unsupervised GraphSAGE methods \citep{hamilton2017inductive}. We also provide results of two supervised approaches, FastGCN \citep{chen2018fastgcn} and Gated Attention Networks (GaAN) \citep{zhang2018gaan}.

\subsection{Model configurations}
\label{models}
Equation Eq. (\ref{gnn}) provides a general formulation of graph neural networks. Several architectures have been proposed for the choice of $AGGREGATE$ and $COMBINE$.  In all our experiments the basic update rule is the mean pooling variant from \citep{hamilton2017inductive}.
\begin{equation}
    h_{u}^{(l)} \leftarrow \left(W^{(l-1)}\right)^{\top} \cdot MEAN(\{h_{u}^{(l-1)}\}\cup \{h_{v}^{(l-1)}, \forall v \in \mathcal{N}(u)\}),
\end{equation}
where the $MEAN$ operator is the element-wise mean of all vectors in ($\{h_{u}^{(l-1)}\}\cup \{h_{v}^{(l-1)}, \forall v \in \mathcal{N}(u)\})$, and $W^{(0)} \in \mathbb{R}^{P\times P'}$ and $W^{(l-1)} \in \mathbb{R}^{P'\times P'}$, for $l>1$, are learnable linear transformations. 
\par
All GNN aggregation operations are computed in parallel resulting in a matrix representation as follows:
\begin{equation}
    \label{GCNlayer}
    H^{(l)}=\hat{A}H^{(l-1)}W^{(l-1)}
\end{equation}
where $H^{(l)}=[h_{1}^{(l)},h_{2}^{(l)},\dots,h_{N}^{(l)}]^{\top}$ is the matrix of nodes' hidden feature vectors at the $l-$th layer and $\hat{A}=\check{D}^{-1}\check{A}$ is the normalized version of the adjacency matrix with added self-loop $\check{A}=A +I_{N}$ with $\tilde{D}$ being its diagonal degree matrix, i.e.  $\check{D}_{ii}=\sum_{j}\check{A}_{ij}$.

\paragraph{Transductive learning} For Citeseer and pubmed, we use a one layer GNN as defined in \ref{GCNlayer}, to which we apply an exponential linear unit (ELU) as an activation function \citep{clevert2015fast}. For Cora, our encoder is a two-layer GNN: 
\begin{equation}
    f(X,A)=\hat{A}\sigma(\hat{A}XW^{(0)})W^{(1)}
\end{equation}
where $\sigma$ is an  exponential linear unit, and $f(X,A)$ is the concatenation of all nodes' embeddings. In each layer, we compute $P'=512$ features resulting in a node embedding size of $512$.
\par
The normalized temperature-scaled cross entropy loss benefits from training on large batches \citep{chen2020simple}.
For the three citation datasets, we compute the contrastive loss function across all the nodes of the graph (i.e. the size of the minibatch is equal to the number of nodes of the graph).
\paragraph{Inductive learning} For both inductive learning setups on large graphs and on multiple graphs , we use a three-layer mean-pooling encoder with residual units as follows:
\begin{equation}
    H^{(1)} = \sigma(\hat{A}XW^{(0)}_{1}+XW_{2}^{(0)})
\end{equation}
\begin{equation}
    H^{(2)} =\sigma(\hat{A}H^{(1)}W^{(1)}_{1}+H^{(1)}W_{2}^{(1)})
\end{equation}
\begin{equation}
    f(X,A) = \hat{A}H^{(2)}W^{(2)}_{1}+H^{(2)}W_{2}^{(2)}
\end{equation}
We set the hidden layers and the embedding size to $P'=512$ and apply ELU as an activation function.
\par 
For the multiple-graph setting, we sample one graph at a time from the training set and consider all nodes in the graph as a minibatch to train the contrastive loss function. For the inductive learning on large graphs, the scale of the dataset makes it impossible to fit into GPU memory. We therefore adopt the sub-sampling strategy of \citep{hamilton2017inductive}. We first select a minibatch of nodes and construct a $L$-hop neighborhood subgraph centered at each of them by sampling a fixed size neighborhood. We sample 10 nodes in each of the three levels resulting in $1+10+100+1000 = 1111$ neighboring nodes for each node in the minibatch. We further apply our stochastic transformation approach to each of the subgraphs and compute the contrastive loss to all pairs of the obtained positive examples.
\par
We use Pytorch \citep{NEURIPS2019_9015} and the Pytorch Geometric \citep{fey2019fast} libraries to implement all our experiments. We initialize all models using Glorot initialization \citep{glorot2010understanding} and trained them to minimize the contrastive loss provided in equation Eq. (\ref{eq:loss2}) using the Adam optimizer \citep{kingma2014adam} with an initial learning rate of 0.001. We tune the weight decay in $\{0.001,0.01,0.05,0.1,0.15\}$. We further tune the temperature $\tau$ in the loss function in $\{0.1,0.5,0.8,1.0\}$ and the number of epochs in $\{20,50,100,150,200\}$.
\par
To define the stochastic perturbation, we tune the probability of dropping an edge in $[0.05,0.75]$ and the probability of dropping node features in $[0.2,0.8]$. GraphCL is found to be robust to different choices of the perturbation parameters. However, we found that applying high perturbations to node features (i.e. randomly dropping 50\% to 70 \% of input features) and small perturbations of the graph structure (i.e. randomly dropping 10\% to 20\% of edges) results in stronger representations.

\begin{table}[tb]
    
    \centering
    \caption{Classification accuracy on transductive tasks and micro-averaged F1 score on inductive tasks}
    \begin{tabular*}{\textwidth}{@{}lll @{\extracolsep{\fill} }ll@{}}
     \multicolumn{5}{c}{\textbf{Transductive} } \\ \toprule
     &\textbf{Method} & \textbf{Cora} & \textbf{Citeseer} & \textbf{Pubmed} \\ \midrule
     Unsupervised &Raw features  & $47.9 \pm 0.4\%$ & $49.3 \pm 0.2\%$ & $ 69.1 \pm 0.3 \%  $\\
     &DeepWalk \citep{perozzi2014deepwalk} & $67.2\%$ & $43.2\%$ & $65.3\%$  \\
     &DeepWalk + features & $70.7 \pm 0.6\%$  & $51.4 \pm 0.5 \%$  & $74.3 \pm 0.9\%$ \\
     &EP-B \citep{duran2017learning} & $78.1 \pm 1.5\%$ & $71.0\pm1.4\%$ & $79.6 \pm2.1\%$  \\
     &DGI \citep{velivckovic2018deep} & $82.3 \pm 0.6 \%$ & $71.8 \pm 0.7 \%$  & $76.8 \pm 0.6 \% $  \\
     &\textbf{GraphCL} & \textbf{$\bm{83.6 \pm 0.5 \%}$}  & \textbf{$\bm{72.5 \pm 0.7 \%}$} & \textbf{$\bm{79.8 \pm 0.5 \%  }$} \\ \midrule
     Supervised&GCN \citep{kipf2016semi} & $81.5$ & $70.3$ & $79.0$  \\
     &GAT \citep{velivckovic2017graph} & $83.0 \pm 0.3 \%$ & $72.5 \pm 0.7 \%$ & $79.0 \pm 0.3 \% $ \\ \bottomrule
    \end{tabular*}
    \newline
    \vspace*{0.2 cm}
    \newline
    \centering

    \begin{tabular*}{\textwidth}{@{}ll @{\extracolsep{\fill} }ll@{}}
    
     \multicolumn{4}{c}{\textbf{Inductive} } \\ \toprule
     &\textbf{Method} &\textbf{Reddit} &\textbf{PPI}  \\ \midrule
     Unsupervised & Raw features & $0.585$  & $0.422$ \\
     &GraphSage-GCN \citep{hamilton2017inductive} & $0.908$ & $0.465$ \\
     &GraphSage-mean \citep{hamilton2017inductive} & $0.897$ & $0.486$ \\
     &GraphSage-LSTM \citep{hamilton2017inductive} & $0.907$ & $0.482$ \\
     &GraphSage-pool \citep{hamilton2017inductive} & $0.892$ & $0.502$ \\
    & DGI \citep{velivckovic2018deep} & $0.940 \pm 0.001 $ & $0.638 \pm 0.002 $ \\
     &\textbf{GraphCL} & $\bm{0.951 \pm 0.01} $   & $\bm{0.659 \pm 0.006}$ \\ \midrule
    Supervised & FastGCN \citep{chen2018fastgcn} & $0.937$  & $-$ \\
     & GaAN \citep{zhang2018gaan} & $0.958 \pm 0.001$  & $0.969 \pm 0.002$ \\ \bottomrule
    \end{tabular*}
    \label{tab:results}
    
\end{table}
\subsection{Results}
\label{results}
We present the results of evaluating node representations using downstream node multiclass classification tasks in Table \ref{tab:results}. We report average results over 50 runs of training followed by a logistic regression. Specifically, we use the mean classification accuracy on the test nodes for transductive tasks and the micro-averaged F1 score on the (unseen) test nodes for the inductive setting. We report the results of EP-B provided in \citep{duran2017learning} and \citep{velivckovic2017graph}, and also the results provided in \citep{velivckovic2018deep}.

\par 
We show that the proposed GraphCL outperforms the previous state-of-the-art by achieving the best classification accuracy over the three transductive tasks and the best F1 score on inductive tasks. We note that, except for PPI dataset, GraphCL achieves competitive performance with strong supervised baselines without using label information. We assume that by maximizing agreement between representations that share the same information but have independent noise, GraphCL is able to learn representations that benefit from the richness of information in the graph which compensate for the information provided by the labels. 
\section{Formal Analysis}
We now analyse the computational and model complexity of GraphCL and its connection to mutual information maximization.
\subsection{Computational and model complexity}
Let $\mathcal{G=(V,E)}$ be a graph and $N=|\mathcal{V}|$ the total number of nodes in the graph. Moreover, let $L$ be the number of layers, $M$ be the minibatch size and $R$ be the number of neighbors being sampled for each node in the inductive setting. We assume for simplicity that the dimension of the nodes' hidden features is constant and denote it as $P'$. The computational complexity and space complexity of GraphCL depend on the choice of the encoder. We use the same encoder for the two branches (i.e. each of the subgraphs). For the transductive learning setup, the computational and space complexity are linear with respect to the number of nodes and are respectively $\mathcal{O}(LNP'^{2})$ and  $\mathcal{O}(LNP'+KP'^{2})$ . For the inductive learning, we use a sub-sampling strategy to load the graphs into memory, the computational complexity is then $\mathcal{O}(R^{L}NP'^{2})$ and the space complexity is $\mathcal{O}(MR^{L}P'+LP'^{2})$. The computational complexity is linear with respect to the number of nodes. Both the number of layers $L$ and the number of sampled neighbors $R$ are fixed and user-specified. The space complexity is linear with respect to the minibatch size $M$. The sampling strategy sacrifices time efficiency to save memory which is necessary for very large graphs.
\subsection{Connection to mutual information}
The contrastive loss in equation Eq. (\ref{eq:loss2}) has been proposed as a lower bound estimator of mutual information. A formal proof given by \citep{oord2018representation} shows that:
\begin{equation}
    \label{eq:bound}
    I(h_{u,1},h_{u,2})\geq\log(k)-\mathcal{L},
\end{equation}
where $k$ is the number of negative samples and $I(h_{u,1},h_{u,2})$ is the mutual information between $h_{u,1}$ and$h_{u,2}$:
\begin{equation}
    I(h_{u,1},h_{u,2})= \displaystyle \displaystyle \mathop{\mathbb{E}}_{(h_{u,1},h_{u,2})\sim p_{h_{u,1},h_{u,2}}(\cdot)} \log \left[ \frac{p(h_{u,1},h_{u,2})}{p(h_{u,1})p(h_{u,2})} \right]
\end{equation}
where $p(h_{u,1},h_{u,2})$ is the joint distribution of $h_{u,1}$ and $h_{u,2}$, and $p(h_{u,1})$ and $p(h_{u,2})$ are the corresponding marginals.
\par
Therefore, given any $k$,  minimizing the loss function $\mathcal{L}$ also maximizes the lower bound on the mutual information $I(h_{u,1},h_{u,2})$. 
We note however that \citep{mcallester2018formal} show that the bound in equation \ref{eq:bound} can be very weak and \citep{tschannen2019mutual} suggest that contrastive methods' success highly depends on the choice of the encoder, and cannot be solely attributed to the properties of mutual information.

\section{Conclusion}
We introduced GraphCL, a general framework for self-supervised learning of nodes' representations. The key idea of our approach is to maximize agreement between two representations of the same node. The representations are generated by injecting random perturbations to the graph structure and nodes' intrinsic features. We have conducted a number of experiments on both transductive and inductive learning tasks. Experimental results show that GraphCL outperforms state-of-the-art unsupervised baselines on nodes' classification tasks and is competitive with supervised baselines. In the future, we will investigate the potential of our approach in learning graphs' representations that are robust to adversarial attacks on the graph data.

\section*{Broader Impact}
With the increasing proliferation of data sources, supervised labeling becomes intractable, as it must rely on both domain knowledge and investment of human time. Representation learning has had many recent successes, first transforming the problem of interest to a graph structure, where nodes and edges may be heterogeneous and may be associated with unstructured content. A challenge with representation learning is the proper definition and use of negative samples.  A second challenge is that current approaches are good either at transductive or inductive tasks, not both. We present a novel approach that relies on perturbed representations of neighborhood graphs. The perturbation offers two advantages: first it generates the desired positive examples. Second, it offers some robustness to corruptions in the data (missing or spurious links, missing feature data). We demonstrate that our methods work well on a variety of datasets: citation networks, social networks, and protein-protein interaction networks.  Our proposed approach will make machine learning models more robust and, thus, have positive impact on society.

\begin{ack}

\end{ack}

\medskip

\small
\bibliography{mybib}

\end{document}